# Variable noise and dimensionality reduction for sparse Gaussian processes

**Edward Snelson**
Gatsby Computational Neuroscience Unit
University College London
17 Queen Square, London WC1N 3AR, UK
snelson@gatsby.ucl.ac.uk

**Zoubin Ghahramani**
Department of Engineering
University of Cambridge
Cambridge CB2 1PZ, UK
zoubin@eng.cam.ac.uk

## Abstract

The sparse pseudo-input Gaussian process (SPGP) is a new approximation method for speeding up GP regression in the case of a large number of data points $N$. The approximation is controlled by the gradient optimization of a small set of $M$ 'pseudo-inputs', thereby reducing complexity from $\mathcal{O}(N^3)$ to $\mathcal{O}(M^2N)$. One limitation of the SPGP is that this optimization space becomes impractically big for high dimensional data sets. This paper addresses this limitation by performing automatic dimensionality reduction. A projection of the input space to a low dimensional space is learned in a supervised manner, alongside the pseudo-inputs, which now live in this reduced space. The paper also investigates the suitability of the SPGP for modeling data with input-dependent noise. A further extension of the model is made to make it even more powerful in this regard – we learn an uncertainty parameter for each pseudo-input. The combination of sparsity, reduced dimension, and input-dependent noise makes it possible to apply GPs to much larger and more complex data sets than was previously practical. We demonstrate the benefits of these methods on several synthetic and real world problems.

## 1 Introduction

Gaussian process (GP) models are widely used to perform Bayesian non-linear non-parametric regression and classification. However one of their principal limitations is their $\mathcal{O}(N^3)$ scaling for training, where $N$ is the number of data points. There have been many methods proposed in recent years to address this problem, and bring the scaling down to $\mathcal{O}(M^2N)$ where $M \ll N$ [Tresp, 2000, Smola and Bartlett, 2001, Williams and Seeger, 2001, Csató, 2002a,b, Lawrence et al., 2003, Seeger et al., 2003, Seeger, 2003, Quiñonero Candela, 2004].

Recently we developed the sparse pseudo-input Gaussian process (SPGP) for regression and showed improvements over previous sparse GP methods in a number of ways [Snelson and Ghahramani, 2006]. Firstly the approximation used in the SPGP is a more accurate approximation to the full GP than used by e.g. Seeger et al. [2003]. Secondly the approximation is based on a set of 'pseudo-inputs' which are learned by gradient optimization, and are therefore not constrained to lie on the data points. This contrasts with the previous sparse GP methods which rely on iteratively choosing a subset of the data – the active set – on which to base the approximation. Learning the pseudo-inputs with gradients allows a greater accuracy to be achieved and allows hyperparameters to be learned in one joint optimization.

One limitation of the SPGP is that learning the pseudo-inputs becomes impractical for the case of a high dimensional input space. For $M$ pseudo-inputs and a $D$ dimensional input space we have a continuous $M \times D$ dimensional optimization task. In this paper we overcome this limitation by learning a projection of the input space into a lower dimensional space. The pseudo-inputs live in this low dimensional space and hence the optimization problem is much smaller. This can be seen as performing supervised dimensionality reduction. In section 6, on several real regression tasks, we show that the dimensionality reduction leads to great reductions in training time over the standard SPGP for little loss in predictive accuracy.

The extra flexibility afforded by learning the pseudo-inputs means that the SPGP is capable of modeling input-dependent noise (heteroscedasticity). This is something that is very difficult to achieve with a standard GP without resorting to expensive sampling [Goldberg et al., 1998]. In this paper we explore the capabilities of the SPGP for heteroscedastic regression tasks, and we develop a further extension of the model that allows an even greater degree of flexibility in this regard. We do this by learning individual uncertainty parameters for the pseudo-inputs.

The extensions of the SPGP presented in this paper allow GP methods to be applied to a large variety of data sets. We can now deal successfully with a large number of data points, high dimensional input spaces, and variable noise. The desirable properties of the GP are maintained throughout – we can make fully probabilistic predictions with appropriate variances.

## 2 Gaussian processes for regression

In this section we briefly summarize GPs for regression, but see [Rasmussen and Williams, 2006, Williams and Rasmussen, 1996, Rasmussen, 1996, Gibbs, 1997, MacKay, 1998] for more detail. We have a data set $\mathcal{D}$ consisting of $N$ input vectors $\mathbf{X} = \{\mathbf{x}_n\}_{n=1}^N$ of dimension $D$ and corresponding real valued targets $\mathbf{y} = \{y_n\}_{n=1}^N$. We place a zero mean Gaussian process prior on the underlying latent function $f(x)$ that we are trying to model. We therefore have a multivariate Gaussian distribution on any finite subset of latent variables; in particular, at $\mathbf{X}$: $p(\mathbf{f}|\mathbf{X}) = \mathcal{N}(\mathbf{f}|\mathbf{0}, \mathbf{K}_N)$, where $\mathcal{N}(\mathbf{f}|\mathbf{m}, \mathbf{V})$ is a Gaussian distribution with mean $\mathbf{m}$ and covariance $\mathbf{V}$. In a Gaussian process the covariance matrix is constructed from a covariance function, or kernel, $K$ which expresses some prior notion of smoothness of the underlying function: $[\mathbf{K}_N]_{nn'} = K(\mathbf{x}_n, \mathbf{x}_{n'})$. Usually the covariance function depends on a small number of hyperparameters $\boldsymbol{\theta}$, which control these smoothness properties. For our experiments later on we will use the standard stationary squared exponential covariance with 'automatic relevance determination' (ARD) hyperparameters [MacKay, 1998]:

$$K(\mathbf{x}_n, \mathbf{x}_{n'}) = c \exp\left[-\tfrac{1}{2}\sum_{d=1}^D b_d(x_n^{(d)} - x_{n'}^{(d)})^2\right], \quad (1)$$

where $\boldsymbol{\theta} = \{c, \mathbf{b}\}$ and $\mathbf{b} = (b_1, \ldots, b_D)$.

In standard GP regression we also assume a Gaussian noise model or likelihood $p(\mathbf{y}|\mathbf{f}) = \mathcal{N}(\mathbf{y}|\mathbf{f}, \sigma^2 \mathbf{I})$. Integrating out the latent function values we obtain the marginal likelihood:

$$p(\mathbf{y}|\mathbf{X}, \boldsymbol{\Theta}) = \mathcal{N}(\mathbf{y}|\mathbf{0}, \mathbf{K}_N + \sigma^2 \mathbf{I}), \quad (2)$$

which is typically used to train the GP by finding a (local) maximum with respect to the hyperparameters $\boldsymbol{\Theta} = \{\boldsymbol{\theta}, \sigma^2\}$.

Prediction is made by considering a new input point $\mathbf{x}_*$ and conditioning on the observed data and hyperparameters. The distribution of the target value at the new point is then:

$$p(y|\mathbf{x}_*, \mathcal{D}, \boldsymbol{\Theta}) = \mathcal{N}(y|\mu_*, \sigma_*^2), \quad (3)$$
$$\mu_* = \mathbf{K}_{*N}(\mathbf{K}_N + \sigma^2 \mathbf{I})^{-1} \mathbf{y}$$
$$\sigma_*^2 = K_{**} - \mathbf{K}_{*N}(\mathbf{K}_N + \sigma^2 \mathbf{I})^{-1} \mathbf{K}_{N*} + \sigma^2,$$

where $[\mathbf{K}_{*N}]_n = K(\mathbf{x}_*, \mathbf{x}_n)$ and $K_{**} = K(\mathbf{x}_*, \mathbf{x}_*)$. The GP is a non-parametric model, because the training data are explicitly required at test time in order to construct the predictive distribution, as is clear from the above expression.

GPs are prohibitive for large data sets because training requires $\mathcal{O}(N^3)$ time due to the inversion of the covariance matrix. Once the inversion is done, prediction is $\mathcal{O}(N)$ for the predictive mean and $\mathcal{O}(N^2)$ for the predictive variance per new test case.

## 3 Sparse pseudo-input Gaussian processes

In this section we review the SPGP, but omit its derivation as an approximation to a GP, for which we refer back to the original paper [Snelson and Ghahramani, 2006]. Quiñonero Candela and Rasmussen [2005] also provide a review paper which assesses the relationship between various sparse GP approximations including the SPGP. The SPGP approximation is based on a set of $M$ pseudo-inputs $\bar{\mathbf{X}} = \{\bar{\mathbf{x}}_m\}_{m=1}^M$. We call these *pseudo*-inputs because they are not a subset of the data inputs, but rather parameters to be learned. The pseudo-inputs can be considered to parameterize an approximation to the GP covariance function (1). Leaving aside its derivation, the SPGP covariance function takes on the following form:

$$K^{\text{SPGP}}(\mathbf{x}_n, \mathbf{x}_{n'}) = \mathbf{K}_{nM} \mathbf{K}_M^{-1} \mathbf{K}_{Mn'} + \lambda_n \delta_{nn'}, \quad (4)$$
$$\lambda_n = K_{nn} - \mathbf{K}_{nM} \mathbf{K}_M^{-1} \mathbf{K}_{Mn}.$$

Here $\mathbf{K}_{nM}$ has as its elements $K(\mathbf{x}_n, \bar{\mathbf{x}}_m)$, the covariance between a data point and a pseudo-input. $\mathbf{K}_M$ has as its elements $K(\bar{\mathbf{x}}_m, \bar{\mathbf{x}}_{m'})$, the covariance of the pseudo-inputs themselves. Notice that even though the underlying GP covariance (1) is stationary, the SPGP covariance is a more complicated non-stationary quantity due to the influence of the particular locations of the pseudo-inputs.

The SPGP covariance *matrix* is formed from (4): $[\mathbf{K}_N^{\text{SPGP}}]_{nn'} = K^{\text{SPGP}}(\mathbf{x}_n, \mathbf{x}_{n'})$. The marginal likelihood can then be constructed analogous to (2):

$$p(\mathbf{y}|\mathbf{X}, \bar{\mathbf{X}}, \boldsymbol{\Theta}) = \mathcal{N}(\mathbf{y}|\mathbf{0}, \mathbf{K}_N^{\text{SPGP}} + \sigma^2 \mathbf{I}). \quad (5)$$

The marginal likelihood is a function of the hyperparameters $\boldsymbol{\Theta}$ and the pseudo-inputs $\bar{\mathbf{X}}$, and it is used to train the SPGP. The hyperparameters and pseudo-inputs are learned jointly by maximizing the likelihood using gradient ascent. The computational efficiency arises because the covariance $\mathbf{K}_N^{\text{SPGP}}$ consists of a sum of a low rank part and a diagonal part, and can therefore be inverted in $\mathcal{O}(M^2 N)$ rather than $\mathcal{O}(N^3)$ time.

Just as in the standard GP, the predictive distribution can be computed by considering a new point $\mathbf{x}_*$ and conditioning on the data $\mathcal{D}$:

$$p(y|\mathbf{x}_*, \mathcal{D}, \bar{\mathbf{X}}, \mathbf{\Theta}) = \mathcal{N}(y|\mu_*, \sigma_*^2) , \qquad (6)$$
$$\mu_* = \mathbf{K}_{*M} \mathbf{Q}^{-1} \mathbf{K}_{MN} (\mathbf{\Lambda} + \sigma^2 \mathbf{I})^{-1} \mathbf{y}$$
$$\sigma_*^2 = K_{**} - \mathbf{K}_{*M} (\mathbf{K}_M^{-1} - \mathbf{Q}^{-1}) \mathbf{K}_{M*} + \sigma^2 ,$$

where $\mathbf{Q} = \mathbf{K}_M + \mathbf{K}_{MN}(\mathbf{\Lambda} + \sigma^2 \mathbf{I})^{-1} \mathbf{K}_{NM}$ and $\mathbf{\Lambda} = \text{diag}(\boldsymbol{\lambda})$. After $\mathcal{O}(M^2 N)$ precomputation, the predictive mean and variance can be computed in $\mathcal{O}(M)$ and $\mathcal{O}(M^2)$ respectively per test case.

## 4 Dimensionality reduction

The SPGP improves the accuracy of its approximation by adjusting the positions of the pseudo-inputs to fit the data well. However a limitation of this procedure is that whereas the standard GP only had a small number $|\mathbf{\Theta}|$ of parameters to learn, the SPGP has a much larger number: $MD + |\mathbf{\Theta}|$. Whilst we can adjust the number of pseudo-inputs $M$ depending on our time available for computation, if we have a high dimensional $(D)$ input space the optimization is impractically large. In this section we address this problem by learning a low dimensional projection of the input space.

In order to achieve this dimensionality reduction we adapt an idea of Vivarelli and Williams [1999] to the SPGP. They replaced the ARD lengthscale hyperparameters $\mathbf{b}$ in the GP covariance function (1) with a general positive definite matrix $W$, in order to provide a richer covariance structure between dimensions:

$$K(\mathbf{x}_n, \mathbf{x}_{n'}) = c \exp\left[-\tfrac{1}{2}(\mathbf{x}_n - \mathbf{x}_{n'})^\top W(\mathbf{x}_n - \mathbf{x}_{n'})\right] . \quad (7)$$

$W$ need not be totally general – it can be restricted to be low rank by decomposing it as $W = P^\top P$, where $P$ is a $(G \times D)$ matrix and $G < D$. This is clearly exactly equivalent to making a linear low dimensional projection of each data point $\mathbf{x}_n^{\text{new}} = P \mathbf{x}_n$, and has the covariance function:

$$K(\mathbf{x}_n, \mathbf{x}_{n'}) = c \exp\left[-\tfrac{1}{2}\big(P(\mathbf{x}_n - \mathbf{x}_{n'})\big)^\top P(\mathbf{x}_n - \mathbf{x}_{n'})\right] . \quad (8)$$

We use exactly this covariance structure for dimensionality reduction in the SPGP. However the SPGP covariance function (4) is constructed from covariances between data-points and pseudo-inputs $K(\mathbf{x}_n, \bar{\mathbf{x}}_m)$, and from the covariances of the pseudo-inputs themselves $K(\bar{\mathbf{x}}_m, \bar{\mathbf{x}}_{m'})$. The projection means that we only need to consider the pseudo-inputs living in the reduced dimensional $(G)$ space. Finally we therefore use the following covariances:

$$K(\mathbf{x}_n, \bar{\mathbf{x}}_m) = c \exp\left[-\tfrac{1}{2}(P\mathbf{x}_n - \bar{\mathbf{x}}_m)^\top (P\mathbf{x}_n - \bar{\mathbf{x}}_m)\right]$$
$$K(\bar{\mathbf{x}}_m, \bar{\mathbf{x}}_{m'}) = c \exp\left[-\tfrac{1}{2}(\bar{\mathbf{x}}_m - \bar{\mathbf{x}}_{m'})^\top (\bar{\mathbf{x}}_m - \bar{\mathbf{x}}_{m'})\right] ,$$
$$(9)$$

where the $\{\bar{\mathbf{x}}_m\}$ are $G$ dimensional vectors. Note that it is not necessary to introduce extra lengthscale hyperparameters for the pseudo-inputs themselves because they would be redundant. The pseudo-inputs are free to move, and the projection matrix $P$ can scale the real data points arbitrarily to 'bring the data to the pseudo-inputs'.

Setting aside computational issues for the moment it is worth noting that even with $G < D$ the covariance (8) may be more suitable for a particular data set than the standard ARD covariance (1), because it is capable of mixing dimensions together. However this is not our principal motivation. The SPGP with ARD covariance has $MD + D + 2$ parameters to learn, while with dimensionality reduction it has $(M+D)G+2$. Clearly whether this is a smaller optimization space depends on the exact choices for $M$ and $G$, but we will show on real problems in section 6 that $G$ can often be chosen to be very small.

Just to clarify: the training procedure for the dimensionality reduced SPGP (SPGP+DR) is to maximize the marginal likelihood (5) using gradients with respect to the pseudo-inputs $\bar{\mathbf{X}}$, the projection matrix $P$, the size $c$, and the noise $\sigma^2$.[1] The procedure can be considered to perform supervised dimensionality reduction – an ideal linear projection is learned for explaining the target data. This is in contrast to the many unsupervised dimensionality reduction methods available (e.g. PCA), which act on the inputs alone.

## 5 Variable noise

In [Snelson and Ghahramani, 2006] we showed preliminary results on a synthetic data set that suggested that the SPGP is capable of dealing with some forms of input-dependent noise (heteroscedasticity). In section 6 we investigate these capabilities further by testing the SPGP on some real data sets believed to be heteroscedastic in nature. However the SPGP is limited in its power to model variable noise. In this section we propose a further extension to the model that enables a greater variety of data sets to be effectively modeled.

---

[1] The gradient derivations are complicated and tedious, and so are omitted here. However they closely follow Seeger et al. [2003].

The best way to see the limitation of the SPGP for variable noise is to examine Figure 1b, reproduced from [Snelson and Ghahramani, 2006]. Although the SPGP has a single global noise level $\sigma^2$, the predictive variances will only drop to this level in regions close to pseudo-inputs. Away from pseudo-inputs the predictive variance rises to $c + \sigma^2$ because correlations cannot be modeled in these regions. During training, the SPGP can adjust its pseudo-inputs to take advantage of this by-product of the non-stationarity of the sparse covariance function. By shifting all the pseudo-inputs to the left in Figure 1b, the SPGP models the variable noise vastly better than the standard GP does in Figure 1a. However this comes at a price – the correlations towards the right side of Figure 1b cannot be modeled because there are no pseudo-inputs there.

Our proposed extension gets around this problem by introducing extra uncertainties associated with each pseudo-point. We alter the covariance of the pseudo-inputs in the following way:

$$\mathbf{K}_M \to \mathbf{K}_M + \mathrm{diag}(\mathbf{h}) \,, \qquad (10)$$

where $\mathbf{h}$ is a positive vector of uncertainties to be learned. These uncertainties allow the pseudo-inputs to be gradually 'switched off' as the uncertainties are increased. If $h_m = 0$ then that particular pseudo-input behaves exactly as in the SPGP. As $h_m$ grows, that pseudo-input has less influence on the predictive distribution. This means that the pseudo-inputs' role is not 'all or nothing' as it was in the SPGP. A pseudo-input can be partly turned off to allow a larger noise variance in the prediction whilst still modeling correlations in that region. As $h_m \to \infty$, the pseudo-input is totally ignored. We refer to this heteroscedastic extension as the SPGP+HS.

Figure 2 shows sample data drawn from the marginal likelihood of the SPGP+HS model, where the components of $\mathbf{h}$ have been set to three different values. These values are indicated by the sizes of the blue crosses representing the pseudo-inputs – a larger cross means a lower uncertainty. Notice the different noise regimes in the generated data.

To train the model, we follow the same procedure as earlier – we include $\mathbf{h}$ as extra parameters to be learned by gradient based maximum likelihood. We tested this on the synthetic data of Figure 1, and the predictive distribution is shown in Figure 1c. Now the pseudo-inputs do not all have a tendency to move to the left, but rather the right most ones can partly turn themselves off, enabling the correlations present towards the right of the data set to be modeled very well.

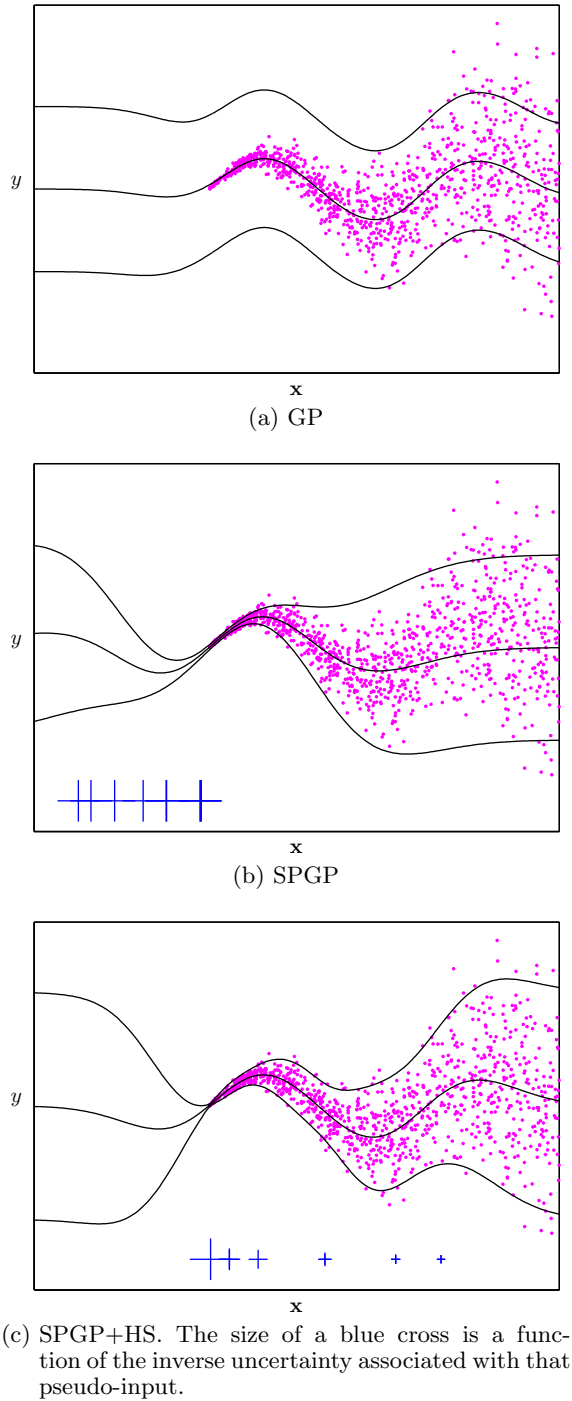

(a) GP

(b) SPGP

(c) SPGP+HS. The size of a blue cross is a function of the inverse uncertainty associated with that pseudo-input.

Figure 1: The predictive distributions after training on a synthetic heteroscedastic data set are shown for the standard GP, SPGP, and SPGP+HS. The data points are the magenta points. The mean prediction and two standard deviation lines are plotted in black. **x** locations of pseudo-inputs are shown as blue crosses (the $y$ positions are not meaningful).

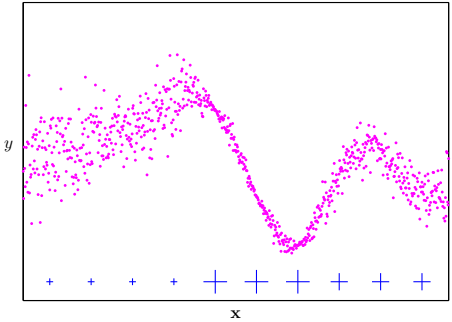

Figure 2: Sample data drawn from the SPGP+HS marginal likelihood for a particular choice of pseudo-input locations (blue crosses), hyperparameters, and pseudo uncertainties. The size of a blue cross is related to the inverse of the uncertainty associated to that pseudo-input.

Our visual intuition is borne out when we look at negative log predictive density (NLPD, with smallest being best) and mean squared error (MSE) scores on a withheld test set for this data set. These are shown below. On NLPD the GP does badly, the SPGP better, but the new method SPGP+HS does best of all, because it models the noise process well. The SPGP is not so good on MSE, because it is forced to sacrifice modeling the correlations on the right side of the data set.

| Method | NLPD | MSE |
| --- | --- | --- |
| GP | 3.09 | 14.16 |
| SPGP | 2.74 | 16.98 |
| SPGP+HS | 2.57 | 14.37 |

## 6 Results

We decided that an ideal test bed for the SPGP and its extensions considered in this paper would be the data sets of the WCCI-2006 Predictive Uncertainty in Environmental Modeling Competition, run by Gavin Cawley[2]. Some of the data sets have a fairly large number of dimensions (>100), and Gavin Cawley suggests that heteroscedastic modeling techniques may well be necessary to perform well on this environmental data. The competition required probabilistic predictions and was to be scored by negative log predictive density (NLPD) on a withheld test set. Unfortunately by the time the competition closed we had only made a submission on one data set (`Temp`), on which we scored first place. However since then, we have experimented further with our methods on the data sets, and Gavin Cawley kindly agreed to evaluate several more submissions on the test sets, which we report here.

Some properties of the data sets we considered are shown in Table 1.[3] Most of the results shown in the following sections are obtained by training on the training set *only* and evaluating on the validation set. This is because the test set targets are not publicly available. These results serve as useful comparisons between our different methods. However some results were obtained by training on the training *and* validation sets, before being sent to Gavin Cawley for evaluation on the withheld test set. With these results we can see how our methods fare against a host of competing algorithms whose performance is shown on the competition web site[2].

[2] http://theoval.sys.uea.ac.uk/competition/

Table 1: Properties of the competition data sets

| Data set | Temp | SO2 | Synthetic |
| --- | --- | --- | --- |
| Dimension ($D$) | 106 | 27 | 1 |
| Training set size | 7117 | 15304 | 256 |
| Validation set size | 3558 | 7652 | 128 |
| Test set size | 3560 | 7652 | 1024 |

### 6.1 `Temp` data set

The targets of the `Temp` data set are maximum daily temperature measurements, and are to be predicted from 106 input variables representing large-scale circulation information. We conducted a series of experiments to see how dimensionality reduction performed, and these are presented in Table 2a. To compare, we ran the standard SPGP with no dimensionality reduction (which took a long time to train). Although the dimensionality reduction did not produce better performance than the standard SPGP, we see that we are able to reduce the dimensions from 106 to just 5 with only a slight loss in accuracy. The main thing to notice is the training and test times, where reducing the dimension to 5 has sped up training and testing by an order of magnitude over the standard SPGP.[4] Clearly some care is needed in selecting the reduced dimension $G$. If it is chosen too small then the representation is not sufficient to explain the targets well, and if it is too large then there are probably too many parameters in the projection $P$ to be fit from the data. Cross-validation is a robust way of selecting $G$.

---

[3] The competition also had a further data set `Precip`, which we have not considered. This is because a histogram of the targets showed a very large spike at exactly zero, which we felt would be best modeled by a hierarchy of a classifier and regressor. The `SO2` data set was somewhat similar but not nearly so extreme, so here we could get away with a $\log(y + a)$ preprocessing transform.

[4] The actual training and test times are affected not just by the number of parameters to be optimized, but also by details of the gradient calculations, where complicated memory/speed trade-offs have to be made. We have tried to implement both versions efficiently.

|         | Validation |         | Time /s |        |
|---------|------------|---------|---------|--------|
| Method  | NLPD       | MSE     | Train   | Test   |
| SPGP    | 0.063      | 0.0714  | 4420    | 0.567  |
| +DR 2   | 0.106(2)   | 0.0754(5)  | 180(10) | 0.043(1) |
| +DR 5   | 0.071(8)   | 0.0711(7)  | 340(10) | 0.061(1) |
| +DR 10  | 0.112(10)  | 0.0739(12) | 610(20) | 0.091(1) |
| +DR 20  | 0.181(5)   | 0.0805(7)  | 1190(50) | 0.148(1) |
| +DR 30  | 0.191(6)   | 0.0818(7)  | 1740(50) | 0.206(3) |
| +HS,DR 5| 0.077(5)   | 0.0728(3)  | 360(10) | 0.062(3) |
| +PCA 5  | 0.283(1)   | 0.1093(1)  | 200(10) | 0.047(2) |

(a) `Temp`. $M = 10$ pseudo-inputs used.

|         | Validation |         | Time /s |        |
|---------|------------|---------|---------|--------|
| Method  | NLPD       | MSE     | Train   | Test   |
| SPGP    | 4.309(2)   | 0.812(1) | 890(40) | 0.723(6) |
| +DR 2   | 4.349(1)   | 0.814(2) | 80(5)   | 0.165(2) |
| +DR 5   | 4.325(1)   | 0.815(4) | 160(5)  | 0.233(1) |
| +DR 10  | 4.323(3)   | 0.809(5) | 290(15) | 0.342(2) |
| +DR 15  | 4.341(3)   | 0.803(6) | 400(10) | 0.458(5) |
| +DR 20  | 4.350(3)   | 0.807(2) | 530(15) | 0.562(4) |
| +HS     | 4.306(1)   | 0.809(2) | 860(30) | 0.714(4) |
| +PCA 5  | 4.395(1)   | 0.855(2) | 170(10) | 0.255(3) |

(b) `SO2`. $M = 20$ pseudo-inputs used.

Table 2: Results showing NLPD and MSE score (smaller is better) on the validation sets of two competition data sets, `Temp` and `SO2`. Times to train on the training set and test on the validation set are also shown. SPGP indicates the standard SPGP, +DR $G$ indicates dimensionality reduction to dimension $G$, +HS indicates the heteroscedastic extension to the SPGP has been used, +PCA $G$ means PCA to dimension $G$ before standard SPGP. Where possible trials were repeated 5 times and standard errors in the means have been reported – numbers in parentheses refer to errors on final digit(s).

Of course a much simpler way of achieving a linear projection of the input space is to do PCA before using the standard SPGP on the smaller dimensional space. In this case the projection is made completely ignoring the target values. The idea behind the SPGP+DR is that the target values should help in choosing the projection in a supervised manner, and that better performance should result. To test this we used PCA to reduce the dimension to 5, before using the SPGP. The results are shown in Table 2a as well. We see that the SPGP+PCA performs significantly worse than the SPGP+DR both on NLPD and MSE scores. The equivalent reduction to 5 dimensions using the SPGP+DR does not cost too much more than the PCA method either, in terms of training or test time.

Our entry to the competition was made by using the SPGP+DR with dimensionality reduction to $G = 5$, and $M = 10$ pseudo-inputs. We trained on the training set and validation sets, and obtained test set NLPD of 0.0349 and MSE of 0.066, which placed us first place on the `Temp` data set on both scores (see the competition web site[2]). This provides justification that the SPGP+DR is a very competitive algorithm, managing to beat other entries from MLPs to Support Vector Regression, and requiring little training and test time.

We then decided to investigate the heteroscedastic capabilities of the SPGP, and the SPGP+HS extension proposed in section 5. Table 2a reports the performance of the SPGP+HS when combined with a dimensionality reduction to $G = 5$. In this case the extension did not perform better than the standard SPGP. However, it could be that either the `Temp` data set is not particularly heteroscedastic, or that the SPGP itself is already doing a good job of modeling the variable noise. To investigate this we trained a standard GP on a small subset of the training data of 1000 points. We compared the performance on the validation set to the SPGP ($M = 10$) trained on the same 1000 points. Since the SPGP is an approximation to the GP, naïvely one would expect it to perform worse. However the SPGP (NLPD 0.16, MSE 0.08) significantly outperformed the GP (NLPD 0.56, MSE 0.11). We suspect that the SPGP does a good job of modeling heteroscedasticity in this data set – something the GP cannot do. The SPGP+HS proposed in section 5 could do no better in this case.

Of course the gradient optimization of the likelihood is a difficult non-convex problem, with many local minima. However the performance seems fairly stable to repeated trials, with relatively low variability.

### 6.2 `SO2` data set

For the `SO2` data set the task is to forecast the concentration of $SO_2$ in an urban environment twenty-four hours in advance, based on current $SO_2$ levels and meteorological conditions. The results presented in Table 2b show a similar story to those on the `Temp` data set. In this case there are a very large number of data points, but a smaller number of dimensions $D = 27$. Although in this case it is perfectly feasible to train the SPGP in a reasonable time without dimensionality reduction, we decided to investigate its effects. Again we find that we can achieve a significant speed up in training and testing for little loss in accuracy. When we compare reducing the dimension to 5 using PCA to using the SPGP+DR, we again find that PCA does not perform well. There is certainly information in the targets which is useful for finding a low dimensional projection. We also tested the SPGP+HS (with no dimensionality reduction), and we see similar, perhaps slightly better, performance than the standard SPGP. We therefore decided to compile a submission using the SPGP+HS, training on the training and validation sets, to submit to Gavin Cawley for evaluation on the test set. We scored an NLPD of 4.28, and MSE of 0.82. Had we managed to submit this entry to the competition before the deadline, we would have been placed second on this data set, again showing the competitiveness of our methods. This time when a GP is compared to the SPGP on a subset of training data of size 1000, the performance is very similar, leading us to suspect that there is not too much to be gained from heteroscedastic methods on this data.

### 6.3 `Synthetic` data set

The final competition data set is a small 1D data set particularly generated to test heteroscedastic methods. Figure 3 shows plots of the data, and the predictive distributions obtained using a GP, a standard SPGP, and the SPGP+HS. These plots show again that the SPGP itself is very capable of modeling certain types of heteroscedasticity. The SPGP+HS creates a very similar predictive distribution, but is able to refine it slightly by using more pseudo-inputs to model the correlations. Both of these look much better than the GP. We sent submissions of all three methods to Gavin Cawley for him to evaluate on the test set. Either the SPGP (NLPD 0.380, MSE 0.571), or the SPGP+HS (NLPD 0.383, MSE 0.562), would have been placed first under NLPD score. In contrast the GP (NLPD 0.860, MSE 0.573) performed poorly on NLPD score as expected. So again we have further evidence that the SPGP can be a very good model for heteroscedastic noise alone. The SPGP+HS extension may improve matters in certain circumstances – here it actually seems to slightly improve MSE over the SPGP, just as we saw for the synthetic data set of section 5.

### 6.4 Motorcycle data set

We finally tested our methods on a data set from Silverman [1985] – data from a simulated motorcycle accident. This is a very small (133 points) 1D data set, which is known for its non-stationarity. We removed 10 random points for testing, trained on the remainder, repeated the procedures 100 times, and the results are shown below. Here we have to report a failure of our methods. The SPGP does not do much better than a standard GP because it cannot deal with this degree of non-stationarity. The SPGP+HS fails completely because it overfits the data badly. The reason for the overfitting is a bad interaction between all the hyper-parameters, where the lengthscale is driven too small, and the pseudo-noise parameters allow the predictive distribution to pinch in on some individual training data points. Essentially, for such a small data set, we have allowed too much flexibility in our covariance function for all the hyperparameters to be fitted using maximum likelihood.

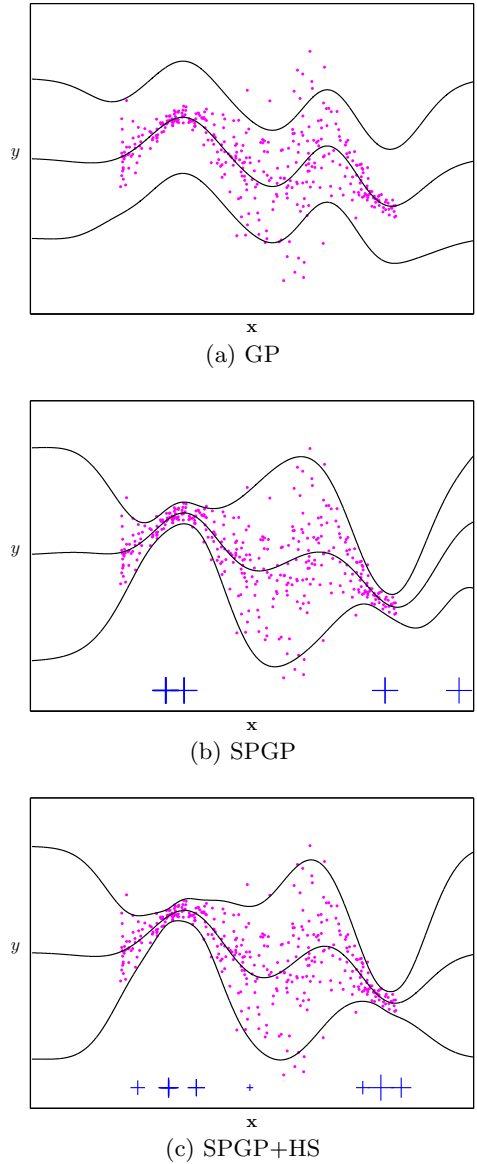

(a) GP

(b) SPGP

(c) SPGP+HS

Figure 3: The predictive distributions on the competition `Synthetic` data set are shown for the standard GP, SPGP, and SPGP+HS.

| Method | NLPD | MSE |
| --- | --- | --- |
| GP | 4.6 | $2.6 \times 10^2$ |
| SPGP | 4.5 | $2.6 \times 10^2$ |
| SPGP+HS | 11.2 | $2.8 \times 10^2$ |

# 7 Conclusions and future work

In this paper we have demonstrated the capabilities of the SPGP and its extensions for modeling data sets with a wide range of properties. The original SPGP could handle data sets with a large number of data points. However it was impractical for data sets with high dimensional input spaces. By learning a linear projection we achieve supervised dimensionality reduction, and greatly speed up the original SPGP for little loss in accuracy. We also have shown the advantage of this supervised dimensionality reduction over the obvious unsupervised linear projection, PCA.

We have also investigated the use of the SPGP for modeling heteroscedastic noise. We find that the original SPGP is a surprisingly good model for heteroscedastic noise, at least in the predictive uncertainty competition data sets. We have also developed an extension of the SPGP more specifically designed for heteroscedastic noise, which although not improving performance on the competition data sets, should provide advantages for some types of problem. However the increase in flexibility to the covariance function can cause overfitting problems for certain data sets, and it is future work to improve the robustness of the method. We could certainly try various forms of regularization and even full Bayesian inference.

There have been a number of previous approaches to developing non-stationary GP models, e.g. Paciorek and Schervish [2004], Higdon et al. [1999]. In contrast to these models, the non-stationarity of the SPGP(+HS) covariance arises directly from the sparse construction. It would be interesting to compare these different approaches further.

The competitiveness of our methods has been demonstrated by our excellent performance on the competition data sets. The scores we achieved on the test sets would have placed us first position on two of the data sets, and second on one. A wide range of other algorithms were competing.

**Acknowledgements**

We would like to thank Carl Rasmussen and Sam Roweis for their useful suggestions, and Gavin Cawley for evaluating some of our competition submissions after the deadline.